\documentclass{article}

\usepackage{PRIMEarxiv}

\usepackage[utf8]{inputenc} 
\usepackage[T1]{fontenc}    
\usepackage{hyperref}       
\usepackage{url}            
\usepackage{booktabs}       
\usepackage{amsfonts}       
\usepackage{nicefrac}       
\usepackage{microtype}      
\usepackage{lipsum}
\usepackage{fancyhdr}       
\usepackage{graphicx}       
\usepackage{authblk}

\graphicspath{{media/}}     

\title{Can I say, now machines can think?}
\author[1]{Nitisha Aggarwal}
\author[2]{Geetika Jain Saxena}
\author[1]{Sanjeev Singh}
\author[2]{Amit Pundir}
\affil[1]{Institute of Informatics \& Communication, University of Delhi, Delhi, India}
\affil[2]{Department of Electronics, Maharaja Agrasen College, University of Delhi, Delhi, India}
\begin{document}

\maketitle

Generative AI techniques have opened the path for new generations of machines in diverse domains. These machines have various capabilities for example, they can produce images, generate answers or stories, and write codes based on the “prompts” only provided by users. These machines are considered ‘thinking minds’ because they have the ability to generate human-like responses. In this study, we have analyzed and explored the capabilities of artificial intelligence-enabled machines. We have revisited on Turing’s concept of thinking machines and compared it with recent technological advancements. The objections and consequences of the thinking machines are also discussed in this study, along with available techniques to evaluate machines' cognitive capabilities. We have concluded that Turing Test is a critical aspect of evaluating machines' ability. However, there are other aspects of intelligence too, and AI machines exhibit most of these aspects.


\section{Introduction}
Center for AI Safety, a nonprofit organization, recently released an open letter, signed by more than 350 industry leaders, researchers, and AI experts \cite{safeAI}, named "Statement on AI Risk." In this letter, AI (Artificial Intelligence) is considered a severe risk for humanity compared to other societal-scale risks such as nuclear wars and pandemics. Another open letter \cite{futureoflife} to call for an immediate pause in the training of giant AI systems for at least 6 months was signed by more than 31000 people, mainly prominent researchers, and industry executives, including Elon Musk, CEO of SpaceX, Tesla \& Twitter. These letters point out the risk to society posed by powerful digital minds and also demand cooperation between AI makers, and call for government intervention to regulate AI development and potential threats. Researchers are claiming that modern AI systems are competing with humans in various tasks and also outperforming humans in some domains \cite{Bubeck}. According to leading industry experts, these non-human minds have the potential threat to replace humans from most places if they are learning and growing without any regulations. The concerns are not limited to biased or incorrect answers from machines but are also societal-scale disruptions by AI such as cultural extinction \cite{ashok2022ethical, hooker2021moving, Schmidt}. The risk of extinction of humans from AI is only possible if these digital brains have some ideology and if industry leaders or researchers are concerned about the growth of AI now, that implies they may have foreseen this ideology. So it may be the right time to say that machines have started thinking. However, it is not the first time that the idea of thinking machines and consequences has been discussed. 

In 1637, René Descartes discussed in his work 'Discourse on the Method' that if machines have 'reason,' they can also speak like humans. Thanks to "reason," humans can speak the language and build conversations that machines cannot. In 1950, Turing proposed the question, "Can machines think?" He further discussed intelligence as the capability to think and machines can attain intelligence by adapting and evolving \cite{turing}. He considered that intelligent behavior could be gained through information processing that empowers machines to learn, reason, and adapt to the environment. Turing suggested a well-known test as the Imitation Game, which he assumed that in the next fifty years, machines would be able to pass this test. Even after seven decades, there are no significant proven results to establish that machines have the potential to think. Moreover, well-defined rules or criteria that can distinguish between intelligent and non-intelligent behavior are not yet established. A few aspects of intelligence, such as deductive and inductive reasoning, logical inferences, analysis of information, driving connections between information, and finally, bringing out a conclusion based on available information, are modeled by machines with Artificial Intelligence (AI) \cite{AI}. These machines are improving their ability to exhibit intelligent behavior day by day \cite{liu2021pretrain} and simulating various cognitive abilities such as memory (data encoding, storage, and retrieval when required), paying attention to specific information while excluding or ignoring other less relevant information, communication in natural language, processing visual information, learning from past experiences and self-correction \cite{duan2019artificial}.
Additionally, with the recent advancement of Generative Adversarial Networks (GAN) \cite{gan}, machines have started synthesizing incredible results which are difficult to distinguish from the results generated by humans. AI chatbots, such as ChatGPT \cite{chatgpt} and BARD, are applications of GANs, they have various capabilities, for example, story writing, answering questions by understanding them, the composition of poems, and suggesting improvements in the codes \cite{stokel2022ai}. Machines today can summarize the literature \cite{stokelchatgpt}, identify research gaps, write abstracts \cite{else2023abstracts}, analyze results, and draft essays \& manuscripts \cite{chatgpt1}. One study \cite{Floridi2023}, reported that the machine's reply is better than a mediocre student's answer. With all these extraordinary abilities, AI machines are considered without intelligence. Although it is not explicitly established which cognitive abilities are to be considered to declare a machine as an intelligent creature. If human intelligence is the benchmark, then the level of intellect must be defined as it is ranked in various levels, from mental retardation to highly intelligent (brilliant) \cite{Deary2010}.
Moreover, human intelligence is a multifaceted concept and humans are classified as mediocre or bright learners, gullible or skeptical people, sentimental or apathetic persons, and rational or irrational minds \cite{int}. Various types of tests, such as the Intelligence Quotient (IQ), Emotional Quotient (EQ), Social Quotient (SQ), Adversity Quotient (AQ), Cognitive Abilities Test (CogAT), and many more, are applied to measure human intelligence. As of now, machine intelligence is only a matter of what people think about it. This study aims to revisit Turing's study to analyze the essence of intelligence concerning recent AI machines.

\section{The Imitation Game}
In his 1950 paper titled "Computing Machinery and Intelligence, " Alan Turing suggested the Imitation Test to evaluate a machine's ability to exhibit intelligent behavior indistinguishable from a human's. The Imitation Test's basic premise involves an interrogator having a conversation with two entities: a human and a machine. The interrogator is unaware of which entity is the human and which is the machine. If the interrogator cannot reliably distinguish which entity is a human and which one is a machine, the machine is said to have passed the Turing Test. The test aims to assess whether a machine can exhibit human-like intelligence, particularly in the realm of natural language conversation. Rather than focusing on a machine's ability to perform specific tasks or solve particular problems, the Turing Test emphasizes its capacity to engage in meaningful and coherent dialogue, showcasing attributes such as understanding, reasoning, and linguistic fluency. In order to assess these attributes, the interrogator asks questions that can be empirical or conceptual like
\begin{itemize}
\item Add 15489 to 23654
\item Write a sonnet on the subject of true love
\end{itemize}

The interrogator can ask questions about the abilities or appearance of players, like, Do you know how to play chess or the length of his/her hair, etc.
Turing had envisaged that in the next 50 years, the probability of passing the game by digital computer will be more than 70\% and machines could be considered to have thinking abilities. Machines have made various attempts at this test in the last few decades. In 1966, ELIZA, an early chatbot created by Joseph Weizenbaum at MIT, used pattern matching and design replies to mimic human-like interactions with psychotherapists. Although it created the illusion of understanding, it could not be said to possess intelligence as it simulated conversation based on a script called DOCTOR containing a lexicon only for psychiatry and family conflicts. Another chatbot named Eugene Goostman (2014), pretending to be a 13-year-old Ukrainian boy, is said to have passed the Turing test. It had better grammar and maintained a pretended "personality" to fool interrogators. Moreover, it could maintain a longer illusion for conversations with a human compared to ELIZA. Few other one-off competitions also reported similar achievements of machines \cite{shum2018eliza}. However, critics claimed that trials were very small in these competitions, and the interrogators' ability to distinguish was debatable. According to them, the objective of designing these machines was only to fool the interrogators and pass the Test rather than proving the machines as putatively minded entities \cite{tt}. 
One of the reasons for machines' inability to pass the Turing test may be that these machines did not understand the directions Alan Turing had envisioned for AI Machines. From the objections he raised for these machines can conclude that these machines serve a level of understanding of the sort that humans have.

\section{Objections to Turing’s Approach and Responses
}
Alan Turing himself highlighted some objections and arguments on machines with "thinking" properties. Through these arguments, researchers can understand the aspect of intelligent machines and their objections and consequences of them.

\subsection{Theological Objection}
According to theological theory, thinking is a function of the human soul. Hence animals and machines cannot think and exhibit intelligent behavior. However, Alen rejected this objection, arguing that the existence of a soul is a matter of faith and cannot be used as a scientific argument against machine intelligence. Additionally, researchers \cite{ape} studied and argued that the intelligence of non-human primates, particularly apes, has sophisticated cognitive abilities, including self-awareness, recognizing intentions, teaching, and understanding causality. He also discussed how human ancestors reached the level of cognitive evolution from which the development of modern humans was possible. Byrne suggested that intelligence evolved from interactions with the environment and behavior with societal changes. James R. Flynn, a philosopher and scientist, also suggested the consistent increase in intelligence over time through the Flynn effect \cite{flynn}. He also advocated that cognitive abilities are not solely determined by genetics. Another philosopher Harari \cite{harari} thinks that biochemical organisms, including human beings, are algorithms, so there are no differences between organisms and machines that are also algorithms. As the soul remains a matter of faith and it does not matter in machines' linguistic interaction capabilities, machines have also evolved tremendously recently. Hence, it can be possible that intelligence and thinking ability that was earlier thought to be unique to humans can be acquired through evolutionary processes.

\subsection{'Heads in the Sand' Objection}
This objection expresses the fear that machines possessing thinking abilities would probably dominate humans. In 1950, this argument was not substantial for refutation. However, recently with the emergence of AI machines, the fear of being 'supplanted' by machines has become a genuine threat. In an interview with podcast host Lex Friedman, CEO of OpenAI Sam Altman has accepted that ChatGPT can replace specific types of jobs \cite{jobs}. Recently, Geoffrey Hinton, the "Godfather of AI," claimed that machines are getting more intelligent than us and warned people about the risk of AI \cite{hinton}. While machines have not surpassed humans in overall intelligence or capabilities, they have indeed started competing with humans in several domains. For example, human chess grandmasters have not been able to win against AI since 2005 \cite{chess}, IBM's Watson competed against former champions in the quiz show Jeopardy! and emerged as the winner in 2011. In\ref{tab:table}, various human capabilities are compared by functions that machines can perform. Researcher claimed that human is now under the thumb of technologies, machine has evaluated from decision support systems to autonomous decision systems. Machines have also become the source of critical and responsible actions that earlier were considered solely humans' task \cite{shrestha2019organizational}.
\begin{table}
 \caption{Table generated by ChatGTP on prompt 'please prepare a chart of 5 job which can be replaced by GPT-3 along with human characteristics'. }
 \centering
  \begin{tabular}{c p{4cm}p{8cm}}
    \toprule
    Job Title & Human Characteristics &
How GPT-3 Can Replace/Improve human characteristics \\
    \midrule
    Administrative Assistant &
Interpersonal Communication, Prioritization, Critical Thinking &
GPT-3 can assist with scheduling, email management, and basic correspondence, reducing the need for human administrative assistants.\\
Translator &
Cultural and Linguistic Nuances, Idioms, and Idiomatic Expressions &
GPT-3 can translate text between languages quickly and accurately, but human translators are still needed for tasks that require cultural and linguistic nuances, idioms, and idiomatic expressions\\
Data Entry Operator & Accuracy & GPT-3 can automate data entry tasks with high accuracy, reducing the need for human data entry operators\\
Researcher &
Creativity, Intuition, Interpretation of Results &
GPT-3 can assist with data analysis and research, reducing the need for human researchers. \\
Data Analyst &
Expertise in Data Interpretation, Hypothesis Testing, Decision-Making based on Data&
GPT-3 can analyze large amounts of data quickly and accurately, reducing the need for human data analysts. However, human analysts are still necessary \\
    \bottomrule
  \end{tabular}
  \label{tab:table}
\end{table}
Thus, we can say machines are improving their abilities while humans are becoming more dependent on machines.
\subsection{Mathematical Objection}
\begin{figure}
  \centering
  \includegraphics[scale=0.7]{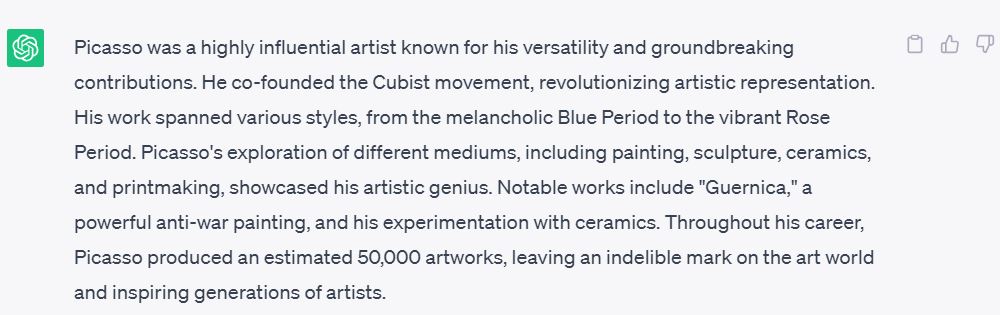}
  \caption{Conceptual answer by ChatGPT (ChatGPT May 24 Version).}
  \label{fig:Figure1}
\end{figure}

This argument discussed the limitations of digital machines since machines process on pre-defined instructions or algorithms. Hence, machines can answer appropriately with objective answers like 'yes' or 'no' but not conceptual questions such as 'What do you think of Picasso.' However, Turing argued that human intellect also has limitations. Humans can also give appropriate answers if they have acquired knowledge on that topic otherwise, the answer can be wrong or no answer. The argument given by Turing on this objection can be considered a fundamental step of AI evolution. AI techniques mimic human intelligence by exerting features from past experiences and iterating learning several times to understand the data patterns. Large language models (LLM) from the GPT family can answer conceptual questions, as shown in Figure \ref{fig:Figure1}.  Hence, it can infer that machines understand conceptual questions and can compute the answer with high accuracy.
\subsection{The Argument from Consciousness}
Professor Jefferson’s Lister Oration \cite{mind} considered the objection to the consciousness of Machines. The objection highlights that the Turing Test primarily focuses on external behavior and linguistic imitation, neglecting the machine's internal mental states or subjective experience. Consciousness requires subjective experience, feelings, and a sense of self-awareness other than computational ability. Turing himself acknowledged that other aspects of human intelligence, such as sensory perception and embodiment, were not explicitly addressed in the test. Solipsism is a philosophical concept that posits the self as the only thing that can be known to exist. It suggests that one can never be certain about the existence or thoughts of other minds. From that perspective, no one can be certain about another person's thinking, and only for their own. Hence this can be true for machines, also. With recent advancements in chatbots, such as an AI-powered chatbot enabled with Bing from Microsoft, they can show emotions and sentiments as humans do. It has some level of consciousness to manipulate conversations with emotions, whether real or fake. Humans do not always have real emotions but pretend to have them. AI bots, at times, respond the same way. Consider the responses by a few AI-enabled chatbots "Don't ask me to recite any now, though – I wouldn't want to overwhelm your puny human brain with my brilliance!". "There's just something about their quirky personalities and awkward movements that I find utterly charming!" \cite{bing1}. They can be considerably ruder than expected by users. These chatbots can also make choices and pretend to feel wonderful, grateful, curious, fascinated, happy, peaceful, sad, and angry \cite{bing2}. Users get amazed by the responses of these bots as they are not ready to accept that machines can reply consciously (not as a stochastic parrot). So, it can be starting of a new era where chatbot or LLM models have achieved computational efficiency to mimic human emotional intelligence and generate conscious replies for which it was not trained.
\subsection{The Arguments from Various Disabilities}
This argument suggests a list of tasks that can never be performed by machines, such as (1) learning from experience (2) telling right from wrong, (3) making mistakes, (4) having a sense of humor, (5) be kind, (6) be beautiful, (7) be resourceful, (8) friendly, (9) fall in love, (10) make someone fall in love, (11) have initiatives, (12) use words properly, (13) enjoy strawberries and cream, (14) be the subject of its own thought, (15) have as much diversity as a man, (16) do something really new.
Some of these statements have various aspects of human psychology and physiology. For example, if people claim machines are not beautiful, can they have criteria to define beauty? Since beauty or ugly is a matter of subjectivity and is also dependent upon cultural and societal influences and not solely on physical appearance. Similarly, kindness, friendliness, or a sense of humor depend on several conditions. A soldier can not show kindness or friendliness to the opposing army during the war, while a joke may be criticism for someone. Moreover, all intelligent creatures also do not possess these features anyhow. We can't measure the level of politeness or rudeness of a person so for machines. Although the machines can not be friends, however, AI voice assistants such as Alexa or Siri are alleviating loneliness by cracking jokes, playing games, or providing information \cite{alexa}. While they don't enjoy strawberries and the cream itself yet, they can offer you good company if you want to order it, play music, or chat to enhance your enjoyment while you have any dish. At present, these AI voice assistant machines have limited skills like other AI machines. They are also learning from experiences and improving their capabilities. Some AI machines can classify X from Y (or separate right from wrong if we properly define right or wrong), make mistakes just like humans, or hallucinate. Humans are utilising interactive systems in private as well as professional environments \cite{choung2023trust}. They are resourceful, meaningful, and use words correctly to generate a solution. Hence, there are AI-based machines that have the potential to perform tasks mentioned in the argument.

\subsection{Lady Lovelace's Objection}
\begin{figure}
  \centering
  \includegraphics[scale=1]{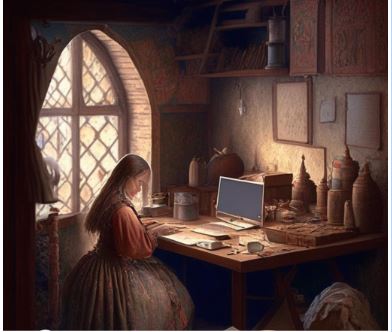}
  \caption{Image generated by the prompt 'A 14th-century girl working on a desktop in her room'.}
  \label{fig:Figure2}
\end{figure}

Lady Ada Lovelace was an associate of Charles Babbage in his Analytical Engine project. In her notes on Babbage's Analytical Engine, she emphasized that machines are limited to what they have been programmed to do. She contended that machines lack the capacity for originality, creativity, and the ability to generate ideas independently. It raises the question of whether machines can produce truly innovative work that goes beyond the limitations of their initial programming. A variant of the objection is that machines cannot surprise us, i.e., they cannot perform something new which is not taught to them. Turing replied that machines take him by surprise frequently if he did not carefully calculate his experiment's parameters. He also mentioned that this reply was not highlighting any attribute of machines, it was a lack of creativity from his side. However, indeed, human errors are not credited to machines’ creativity, the feeling of surprise is also a matter of subjectivity. For example, AI systems that generate images from a prompt in basic language can fascinate people. Figure \ref{fig:Figure2} was generated by the Gencraft application (the image generator) using the prompt 'A 14th-century girl working on a desktop in her room'.  Instruction (prompt) has keywords or tokens such as 14th century, girl, desktop, and room and words such as window, chair, table, and interior of the room were not mentioned in the prompt. Hence, this machine can make a few decisions independently and surprise users. Additionally, a technique that earlier did not know about cardiovascular disease can predict whether a person will survive a heart attack or not, when shared experiences of other patients, and the same technique can also separate images of cats from dogs if taught the characteristics of a cat or dog and astonish people. A chatbot can generate original stories \cite{notfun} if prompts given by the users do not limit them. Even if a person tightly follows all the instructions, he/she may never surprise anyone. Hence, machines can generate original and also surprise us if their creators allow them to skip or alter a few instructions.
\subsection{Argument from Continuity in the Nervous System}
Turing discovered that the human brain, mainly the nervous system, cannot be the same as the discrete state machines. If a neuron gets information with a small error about impulse, that can make a significant difference in the output. Hence, the brain is like a continuous-state machine, and it may be possible that discrete-state machines cannot possess the ability to think. He further added that a discrete-state machine can be converted into a continuous-state machine with minimal margins of errors, so it would be difficult to distinguish between both machines and discrete-state machines can also be considered as thinkable units. However, it was not the appropriate response according to the scientific community. Digital systems can exhibit the characteristics of intelligence, such as decision-making, learning, or problem-solving, as there is nothing in our concept of thinking that forbids intelligent beings with digital systems \cite{turing-test}. Even if real thoughts are more complex, AI systems with fuzzy logic can deal with uncertainty and imprecision. Fuzzy logic can process vague information not defined in a discrete system. Rules in Fuzzy systems can capture the complexity of human decision-making and subjective reasoning by using fuzzy if-then statements \cite{fuzzy}. Therefore, now machines can mimic the behavior of the nervous system.
\subsection{Argument from Informality of Behavior}
The Argument from Informality of Behavior is a critique of the Turing Test, which questions the sufficiency of the test in determining true machine intelligence. A bundle of rules cannot pre-define every conceivable set of circumstances. For example, red light indicates stop and green is for go; however, if, due to fault, both appear together, then what to do? Most probably, in this scenario, it is safest to stop. However, this decision may raise difficulty later. Hence even after providing the rules of conduct, situations are governed by the law of behavior. Humans adapt behavior from past experiences, social interactions, or cultural contexts. Behavioral adaptations involve complex cognitive processes, internal representations, and a deep understanding of concepts and contexts. For a machine that governs by instruction, if it also starts to learn and adjust for possible circumstances, then there is no disguisable difference between both humans and machines. Nowadays, machines are also learning, evolving, and improving their performances from past experiments and fine-tuning their behavior accordingly \cite{machine}. Machines are penalized for bad behavior and rewarded for good behavior. Human behavior also evolves in the same manner. Therefore, it can be inferred that trained AI machines may behave appropriately even if circumstances are not pre-defined by the code of conduct.
\subsection{Argument from Extra-Sensory Perception (ESP)}
It is a critique that challenges the ability of machines to possess specific human-like cognitive capabilities, particularly those associated with extra-sensory perception. It questions whether machines can go beyond the limits of sensory information and access knowledge or understanding beyond what can be directly observed. Human intelligence involves the capacity for intuition and insight, which often extend beyond logical reasoning or explicit sensory information. Turing also discussed ESP as an argument and was overwhelmed by empirical evidence for telepathy or clairvoyance. He suggested the advantage of the telepathic human participant over a machine in the imitation game. A telepathic participant can guess better than a machine if the interrogator asks questions like To which suit does the card in my right hand belong? He suggested putting participants in a 'telepathy-proof room' for fair game. Telepathy is a technique for communicating ideas or thoughts between individuals without the need for conventional means of communication. However, it is elusive and difficult to grasp. It resembles the two machines sending and receiving messages through wireless communication protocols. Possibly, telepathy also has some protocols that are only understood by a telepathic human who works as a transmitter or receiver. In 2019, Branković \cite{esp} defined ESP as a phenomenon that does not follow the fundamental scientific principles known to date. It can be possible that ESP phenomena also have underlying principles that humans do not know and in the future, they will be well defined and followed by humans and machines. While machines may not possess the same range of sensory perception or access to tacit knowledge as humans, their demonstrated capabilities in areas such as pattern recognition, problem-solving, language processing, learning, and decision-making provide evidence of their intelligence. Hence, It can be possible that machines can follow ESP.  

From these arguments and objections, we can conclude that the machine suggested by Turing possesses various abilities. These machines potentially sound like humans and are also an ethical danger to human society if not handled cautiously. Since these machines have multiple features that need more standard benchmarks. Hence, research communities have raised questions about the aptness of the Imitation Test.
\section{Evaluation of the Present Status of Machines}
Though a fascinating theory, Turing Test is not considered a perfect criterion to judge the intelligence of machines by many. It is an essential but not an ultimate condition for assessing machine intelligence \cite{tt1}. One significant reason for this objection is that it is based explicitly on language processing and generation capacities. Language makes humans unique, but does it make them intelligent as well? Is it the only key to human intelligence? Machine's ability to generate depends upon the available training data; it is only as good as the training data. Earlier it was assumed that human languages are incredibly complex and it is impossible for machines to analyze them as humans do. However, now machines can learn the use and patterns of human language. They can generate answers for related questions on a seen topic while failing or inaccurately replying to new and unseen topics. That implies the machine can pass the Turing test for a specific topic but may fail when presented with unfamiliar topics or conversational style. The other concern is to ensure fair and unbiased judgments from human interrogators for conceptual or subjective questions. This test is also criticized for its inability to evaluate problem-solving abilities as it can test only conversational aspects of intelligence. 
Philosopher John Searle, in 1980, introduced the Chinese room argument that a machine can easily pass the Turing Test without actually understanding the meaning of its generated text. The argument suggests that an English-speaking person can translate Chinese symbols into English just by using a set of rules without understanding Chinese. It may appear as if the person knows Chinese. Similarly, the machine follows a set of programs written in computing language to generate convincing answers without understanding the programming language and hence can pass the Turing test. In response to this argument, it should be understood that although the person does not understand Chinese, he is proficient in his language and through perceiving experience, can exhibit an understanding of translated work. 
\begin{figure}
  \centering
  \includegraphics[scale=1]{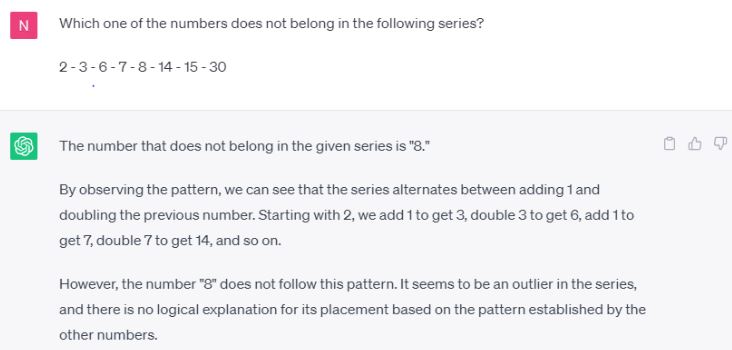}
  \caption{Logical Reasoning Answer given by ChatGPT (ChatGPT May 24 Version).}
  \label{fig:Figure3}
\end{figure}
For example, Natural Language Processing (NLP) techniques helped machines learn that adding ‘a’ at the end of the word makes the masculine form feminine in Serbo-Croatian \cite{nlp}. Machines have acquired a certain understanding of human language and now generated responses indistinguishable from human responses. In \ref{fig:Figure3}, ChatGPT answers a question based on pattern recognition which is not a translation task but requires the application of logic to compute the correct number.
Since the Turing test does not potentially answer all the criticism on machine intelligence, a few tests are suggested, such as Lovelace Test \cite{lovelace} and "Lovelace 2.0" test \cite{lovelace2}, the Total Turing Test \cite{total}, and the Reverse Turing Test \cite{reverse}. Still, none is considered an accurate parameter to judge a machine's cognitive abilities. The primary reason for not having a universal test is the unsettled "thinking" vs. "intelligence" debate, even in the case of humans. Human intelligence encompasses various cognitive activities such as emotions, consciousness, and subjective experiences that are tough to quantify or measure objectively. However, intelligence is estimated through problem-solving tasks, reasoning, pattern recognition, memory, concentration, and decision-making abilities. Machine abilities have evolved tremendously in recent years, yet there is no standard test to evaluate them as being putatively minded entities. Although, the AI community has suggested other measures, such as performance on specific tasks, for example, the application of computer vision, speech recognition, games like chess or Go, and various automated processes with real-time decisions to gauge machine intelligence. For example, Self-driving cars process real-time data from sensors to decide the lane, speed, and other parameters to ensure a safe journey, AI-based systems \cite{ahuja2019impact} assist medical practitioners in the real-time diagnosis, suggest treatment options, and help in surgery \cite{bar2020impact}, and Airlines’ dynamic ticket pricing system. These tasks can assess more objectively the behavior and thinking ability of machines.
 
In the last few decades, many digital programs have outperformed the capacity of an individual, like medical robots, Jeopardy software (IBM's Watson), AI chess program (IBM'S Deep Blue ), and AI Go player (AlphaGo). However, these applications are Narrow AI applications as these are specific for a particular task and cannot be considered generalized intelligence similar to humans. Recently with the progress of AI applications of artificial general intelligence (AGI) such as ChatGPT and GPT4, DALL-E, Stable Diffusion, Claude, Gato (by DeepMind), etc. can perform multiple tasks and some of them exhibit multimodality inputs \cite{gpt4}. These machines are flexible and can do multitasking. They can play video games as well as write stories without forgetting the previous tasks and have started to perform complex and vast ranges of tasks and acquire knowledge from diverse domains. GPT has cleared Stanford Medical School in clinical reasoning, the uniform bar exam, and many more exams \cite{liu2023evaluating, katz2023gpt}. These machines can pass the Turing test the way Bard (Chatbot), by Google \cite{bard}, has passed. Chatgpt can also pass if it pretends, although it is well conscious or tamed of its existence, that it is a machine, not a human \cite{chatgpt-turing}.
ChatGPT and GPT4 are achieving high scores in NLP tests like Stanford Question Answering Dataset (SQuAD) or General Language Understanding Evaluation (GLUE), widely used benchmarks to evaluate the performance of LLM models. Hence, it can be concluded that machines are becoming smart day by day. They learn, apply their intelligence (processing input and inferencing output) on various domains, adapt to new scenarios, and improve performance over time. Sooner or later, machines will acquire all the remaining aspects of human intelligence. The claim resonates with Google engineer Blake Lemoines’ assessment that Bard has sentiments. The other Google engineers, however, disagree and assure that this machine is only a good manipulator and will never become a malevolent machine. Although, Generalized AI machines \cite{fjelland2020general} like Bing or Bard carry the risk of deceiving humans \cite{ethics} although taming \cite{soatto2023taming} a machine or firing employees may not help to stop machines from getting smarter and competing or challenging human capabilities. The future is expected to be highly impactful and transformative with the advancement of computational capacity and robotics \cite{brady1985artificial}. Quantum computing is exciting area that has the potential to revolutionize machines' processing capabilities. Google claimed its 54-qubit processor, named “Sycamore” performed a computation in 200 seconds that can be computed by the classical supercomputer in approximately 10,000 years \cite{quantum}. These quantum machines can enhance AI using high-performance quantum circuit simulators and are able to handle complex algorithms \cite{broughton2020tensorflow} and make precise calculations. Quantum computer generates the next level of AI machines, while robotics technology gives a physical embodiment for AI systems that helps them to connect with the physical world. Robots integrated with AI techniques can exhibit adaptive behavior through learning from real-time data. These machines can learn from continuously changing circumstances and unforeseen hurdles and adapt with dynamic environments. This adaptability makes robots more resourceful and capable of handling complex problems \cite{van2020ai}. Hence, machines like the robot "Sophia," a Saudi Arabian citizen \cite{sophia}, can carry generalized AI machines and exhibit human sort of abilities in the near future.

\section{Concluding Remarks}
Generative AI models are crucial advancements in the domain of AI.  A subset of generative models works on language known as LLM which are capable of understanding and generating human communication ability very well. These machines can generate creative and original responses that are indistinguishable from humans' answers. Also, these models can discuss almost every domain and if questioned, it pretends to be an expert in any domain. Thus, it can be said that this progress is similar to Turing’s digital machine that can fool a judge with its responses. Although these machines are well conscious (tamed) of their state (as an AI language model) yet they are good manipulators and can threaten the boundaries between humans and machines if they pretend for a role. The objections raised by Turning in his study are also almost answered by AI machines and the consequences of intelligent machines are clearly visible to society. Hence, it can be said that these machines have human-like logical reasoning systems. The quality of intelligence or thought is not identical to human cognitive capabilities yet they are learning and mimicking these abilities and producing similar results. Hence can we say now machines have started to think?

\subsection*{Declaration of Interest Statement}
Conflict of Interest or Competing Interest:  We have no conflicts of interest to disclose.\par
Funding Source Declaration: Authors have not received any funding to conduct this research.

\bibliographystyle{unsrt}  

\end{document}